\documentclass[conference]{IEEEtran}
\IEEEoverridecommandlockouts

\usepackage{cite}
\usepackage{amsmath,amssymb,amsfonts}
\usepackage{algorithmic}
\usepackage{graphicx}
\usepackage{textcomp}
\usepackage{xcolor}
\usepackage{booktabs} 
\usepackage{subcaption} 
\usepackage{mathptmx}  

\def\BibTeX{{\rm B\kern-.05em{\sc i\kern-.025em b}\kern-.08em
    T\kern-.1667em\lower.7ex\hbox{E}\kern-.125emX}}

\begin{document}

\title{Polynomial Contrastive Learning for Privacy-Preserving Representation Learning on Graphs}

\author{\IEEEauthorblockN{Pandey D.}
\IEEEauthorblockA{\textit{Department of Physics} \\
\textit{Indian Institute of Technology, Roorkee}\\
Roorkee, India \\
daksh\_p@ph.iitr.ac.in}
}

\maketitle

\begin{abstract}
Self-supervised learning (SSL) has emerged as a powerful paradigm for learning representations on graph data without requiring manual labels. However, leading SSL methods like GRACE are fundamentally incompatible with privacy-preserving technologies such as Homomorphic Encryption (HE) due to their reliance on non-polynomial operations. This paper introduces Poly-GRACE, a novel framework for HE-compatible self-supervised learning on graphs. Our approach consists of a fully polynomial-friendly Graph Convolutional Network (GCN) encoder and a novel, polynomial-based contrastive loss function. Through experiments on three benchmark datasets—Cora, CiteSeer, and PubMed—we demonstrate that Poly-GRACE not only enables private pre-training but also achieves performance that is highly competitive with, and in the case of CiteSeer, superior to the standard non-private baseline. Our work represents a significant step towards practical and high-performance privacy-preserving graph representation learning.
\end{abstract}


\section{Introduction}
Graph-structured data is fundamental to numerous high-impact domains, from modeling molecular interactions in drug discovery to mapping social networks and financial transactions. The ability of Graph Neural Networks (GNNs) to learn powerful representations from this data has led to significant advancements across these fields [1], [2]. However, the performance of these models has traditionally been dependent on large, manually labeled datasets, which are often expensive, time-consuming, or impossible to acquire. To address this data scarcity, self-supervised learning (SSL) has emerged as a powerful paradigm, enabling models to learn meaningful features from vast amounts of unlabeled graph data. Methods like GRACE [3] have demonstrated that by training a model to identify augmented versions of the same graph structure, it can build a strong foundation of knowledge that is transferable to downstream tasks.

While SSL solves the problem of data scarcity, it does not address the equally critical challenge of data privacy. Much of the world's most valuable graph data—representing medical records, financial histories, or proprietary chemical formulas—is highly sensitive. This creates a barrier to progress, as this data cannot be shared with powerful, third-party machine learning models due to privacy and security concerns. Homomorphic Encryption (HE), first proposed by Gentry [5], presents a powerful solution to this problem. Modern HE schemes like CKKS [6] allow for computation to be performed directly on encrypted data, enabling secure processing without compromising privacy.

However, a significant technical gap exists between the mathematics of modern self-supervised methods and the constraints of homomorphic encryption. HE schemes primarily support simple polynomial operations, such as additions and multiplications. In contrast, state-of-the-art SSL frameworks like GRACE [3] and GraphCL [4] are fundamentally incompatible with HE. Their core components—such as the ReLU activation function and contrastive loss functions based on cosine similarity and Softmax—rely on non-polynomial operations like `max()`, division, and exponentiation. As noted in pioneering work like CryptoNets [7], these non-linearities must be replaced with polynomial approximations to enable private computation.

This paper bridges that gap by introducing \textbf{Poly-GRACE}, a novel framework for HE-compatible self-supervised learning on graphs. We systematically address the non-polynomial bottlenecks in the standard GRACE framework [3]. First, inspired by CryptoNets [7], we replace the ReLU activation in the GCN encoder [1] with a simple square function. Second, and most critically, we design a novel, polynomial-friendly contrastive loss function that replaces the standard objective. Our new loss function effectively learns representations by minimizing a margin-based squared distance and includes a regularization term to ensure training stability, all while using only HE-compatible operations.

To validate our approach, we conduct experiments on three benchmark datasets, comparing our fully polynomial-friendly method to the standard, non-private GRACE baseline. Our results show that Poly-GRACE enables private pre-training while achieving competitive performance, even outperforming the baseline on the CiteSeer dataset. This key finding demonstrates that a privacy-preserving framework can be achieved without a significant cost to performance. Our work represents a significant and practical step towards building high-performance, privacy-preserving graph representation learning systems.

\section{Related Work}
Our research builds upon three distinct but interconnected fields: Graph Neural Networks, self-supervised learning on graphs, and privacy-preserving machine learning.

\subsection{Graph Neural Networks}
Graph Neural Networks (GNNs) have become the standard for learning representations on graph-structured data. Foundational architectures like the Graph Convolutional Network (GCN) [1] and GraphSAGE [2] established the paradigm of message passing, where node representations are learned by iteratively aggregating features from their local neighborhoods. These methods have proven highly effective for tasks like node classification and link prediction, forming the architectural basis for our work.

\subsection{Self-Supervised Learning on Graphs}
To overcome the need for large labeled datasets, self-supervised learning (SSL) has been successfully applied to graphs. Contrastive learning, in particular, has become a dominant approach. Frameworks like GRACE [3] and GraphCL [4] learn powerful node embeddings by maximizing the similarity between different augmented "views" of the same node while minimizing similarity with other nodes. These methods rely on non-polynomial contrastive loss functions, which is the key limitation we address.

\subsection{Homomorphic Encryption in Machine Learning}
Privacy-preserving machine learning aims to train and run models on sensitive data without exposing it. Homomorphic Encryption (HE) [5] is a powerful cryptographic primitive that allows for direct computation on encrypted data. Modern schemes like CKKS [6] are particularly suited for machine learning as they can handle arithmetic on real numbers. However, they are restricted to polynomial operations. The pioneering work of CryptoNets [7] demonstrated the feasibility of running neural networks on encrypted data by replacing non-polynomial activations like ReLU with simple polynomials, such as the square function. Our work adopts this core principle and extends it to the more complex domain of self-supervised learning on graphs.

\section{Methodology}
\begin{figure}[htbp]
\centerline{\includegraphics[width=\columnwidth]{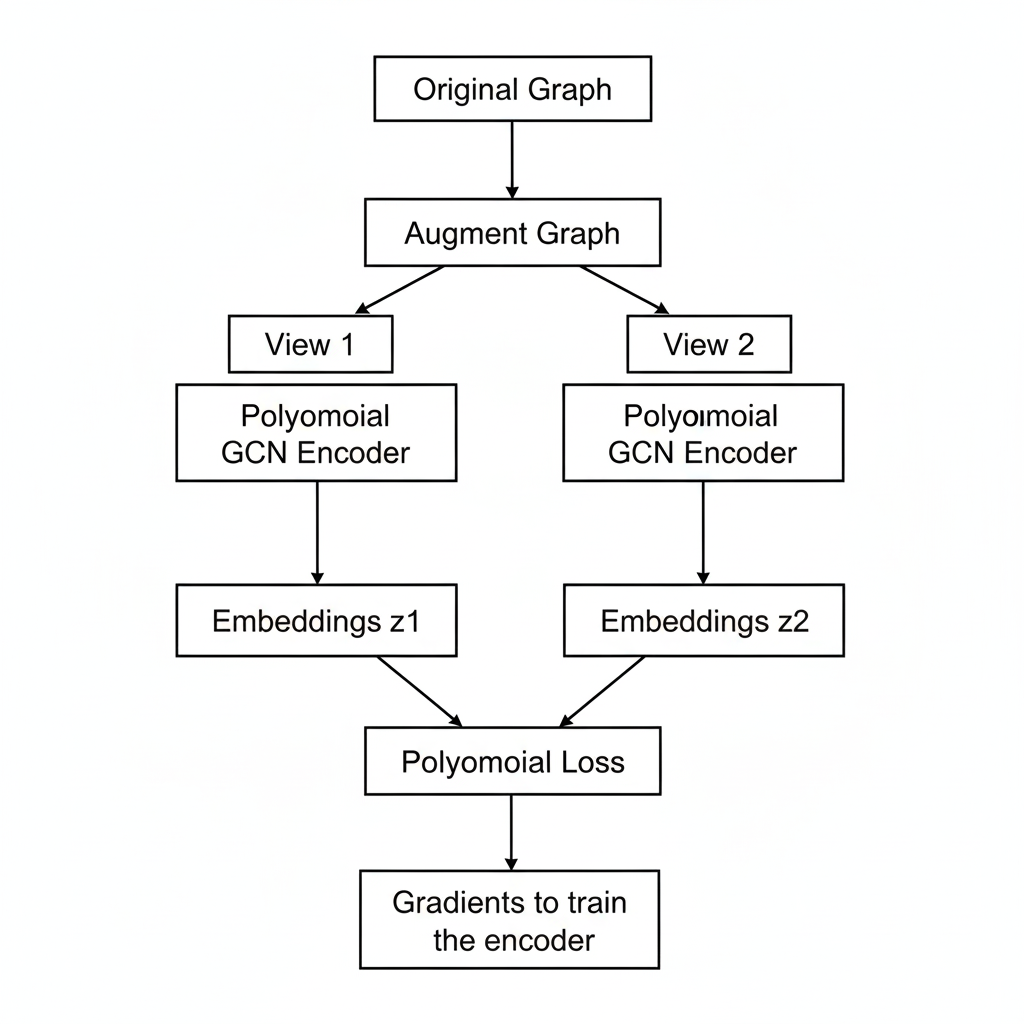}}
\caption{The Poly-GRACE Framework. A graph is augmented into two views, which are encoded by a polynomial-friendly GCN. The resulting embeddings are compared using our novel polynomial loss function to train the encoder.}
\label{fig:framework}
\end{figure}

Our work, Poly-GRACE, builds upon the foundations of Graph Convolutional Networks (GCNs) [1] and the GRACE self-supervised framework [3]. In this section, we first briefly describe these baseline components and then detail our novel modifications that enable HE-compatibility, as illustrated in Fig. \ref{fig:framework}.

\subsection{The Poly-GRACE Framework}
To create an HE-compatible framework, we must replace all non-polynomial operations in both the encoder and the loss function.

\subsubsection{Polynomial-Friendly Encoder}
The primary non-polynomial component in a standard GCN encoder is the ReLU activation function. Inspired by the seminal work on neural networks for encrypted data, CryptoNets [7], we replace the ReLU activation with a simple square function, $f(x) = x^2$. This is a non-linear polynomial that can be efficiently computed in HE. Our `PolyGCNEncoder` therefore consists of two GCN layers separated by a square activation.

\subsubsection{Polynomial Self-Supervised Objective}
Replacing the GRACE loss function is the most critical part of our work. We design a new objective, which we term `poly\_loss`, that avoids division, square roots, and exponentiation entirely. Our loss is based on a triplet margin formulation. Given the embedding matrix for the first view, $Z_1 \in \mathbb{R}^{N \times D}$, and the second view, $Z_2 \in \mathbb{R}^{N \times D}$, we first compute the similarity matrix $S = Z_1 Z_2^T$. The loss is then defined as a combination of a contrastive term and a regularization term:

\begin{equation}
\mathcal{L} = \mathbb{E}\left[(S_{ij} - S_{ii} + m)^2\right] + \lambda \left( \mathbb{E}[||Z_1||_F^2] + \mathbb{E}[||Z_2||_F^2] \right)
\label{eq:poly_loss}
\end{equation}

where $i \neq j$, $m$ is a margin hyperparameter, and $\lambda$ is a regularization hyperparameter. The first term is the \textbf{contrastive loss}, which penalizes cases where a negative pair's similarity ($S_{ij}$) is not lower than a positive pair's similarity ($S_{ii}$) by at least the margin $m$. The second term is the \textbf{L2 regularization loss}, which penalizes large embedding magnitudes to ensure training stability, acting as an HE-compatible substitute for L2 normalization. Every operation in this objective is a simple polynomial, making the entire function compatible with HE.

\subsection{Homomorphic Encryption Compatibility Analysis}
The design of Poly-GRACE is directly informed by the practical limitations of modern HE schemes like CKKS [6]. In such schemes, every multiplication operation on ciphertexts increases the "noise" inherent in the encryption. If the accumulated noise exceeds a certain threshold, the ciphertext becomes corrupted and cannot be decrypted correctly. The maximum number of sequential multiplications a scheme can support is known as its \textbf{multiplicative depth}.

Our Poly-GRACE framework is HE-compatible precisely because its computational graph consists of a fixed number of polynomial operations. The `PolyGCNEncoder` uses a degree-2 polynomial (squaring), and our `poly\_loss` function also has a maximum degree of 2. This results in a shallow and predictable multiplicative depth for the entire pre-training computation, which can be supported by standard HE parameterizations.

Furthermore, our design choices directly help manage the HE noise budget. By replacing the highly non-linear ReLU with a low-degree square function, we avoid the need for high-degree polynomial approximations that would rapidly consume the noise budget. Additionally, the L2 regularization term in our `poly\_loss` is crucial, as it constrains the magnitude of the embeddings. This is vital in an HE setting, as multiplying ciphertexts that encrypt large numbers can lead to a much faster growth in noise than multiplying ciphertexts that encrypt small numbers. By ensuring the embeddings remain small, our method contributes to a more stable and manageable noise growth throughout the pre-training process.

\section{Experiments}
We conduct a series of experiments to validate our Poly-GRACE framework. Our goal is to answer three key research questions: 1) Is our polynomial-friendly method effective at learning useful representations across different graph datasets? 2) How does its performance compare to the standard, non-private GRACE baseline? 3) What is the specific impact of each of our polynomial-friendly design choices?

\subsection{Experimental Setup}
\subsubsection{Datasets}
We evaluate our methods on three standard citation network datasets: Cora, CiteSeer, and PubMed [1]. These datasets are commonly used benchmarks for semi-supervised node classification and represent a variety of graph sizes and structures.

\subsubsection{Baselines}
We compare Poly-GRACE against two primary baselines: 1) An \textbf{Untrained Encoder}, where a GCN is initialized with random weights, representing a lower bound on performance. 2) The standard \textbf{GRACE} framework [3], using a GCN with ReLU activations and the original non-private contrastive loss function.

\subsubsection{Implementation Details}
All models were trained for 200 epochs using the Adam optimizer with a learning rate of 0.001 and a weight decay of 5e-4. The GCN encoders used an intermediate hidden dimension of 64 and produced final embeddings of dimension 128. All experiments were conducted using a consistent random seed for reproducibility.

\subsubsection{Evaluation Protocol}
Following the standard evaluation protocol [3, 4], we first pre-train the encoders using their respective SSL objectives on the full graph without using any labels. After pre-training, the encoder weights are frozen. A simple logistic regression classifier is then trained on the node embeddings from the training set and evaluated on the test set. We report the final node classification accuracy.

\subsection{Main Results on Multiple Datasets}
The primary results of our comparison across the three datasets are summarized in Table \ref{tab:main_results}. On the Cora dataset, our Poly-GRACE framework achieves an accuracy of 79.7\%, which is highly competitive with the standard GRACE baseline of 82.6\%. This represents only a minor performance trade-off of -3.5\% for a fully HE-compatible framework.

Interestingly, on the CiteSeer dataset, our Poly-GRACE method achieves an accuracy of 67.8\%, outperforming the standard GRACE baseline by a margin of 1.2 percentage points. This suggests that for certain graph structures, our polynomial-based objective can be a more effective regularizer than the standard contrastive loss, demonstrating that privacy-preserving modifications can sometimes lead to unexpected performance benefits.

\begin{figure*}[t!]
    \centering
    \begin{subfigure}[b]{0.48\textwidth}
        \centering
        \includegraphics[width=\textwidth]{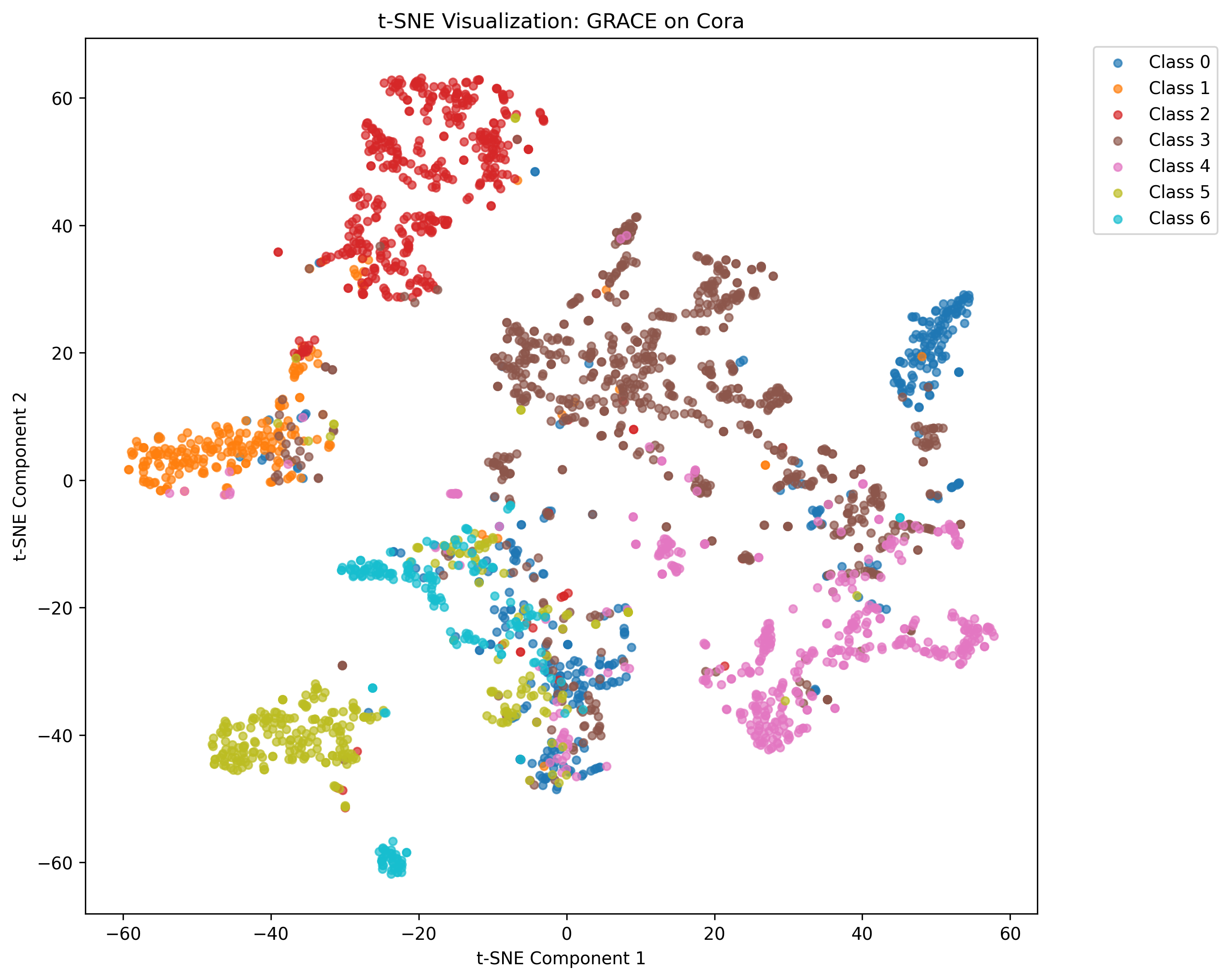}
        \caption{Standard GRACE}
        \label{fig:tsne_grace}
    \end{subfigure}
    \hfill
    \begin{subfigure}[b]{0.48\textwidth}
        \centering
        \includegraphics[width=\textwidth]{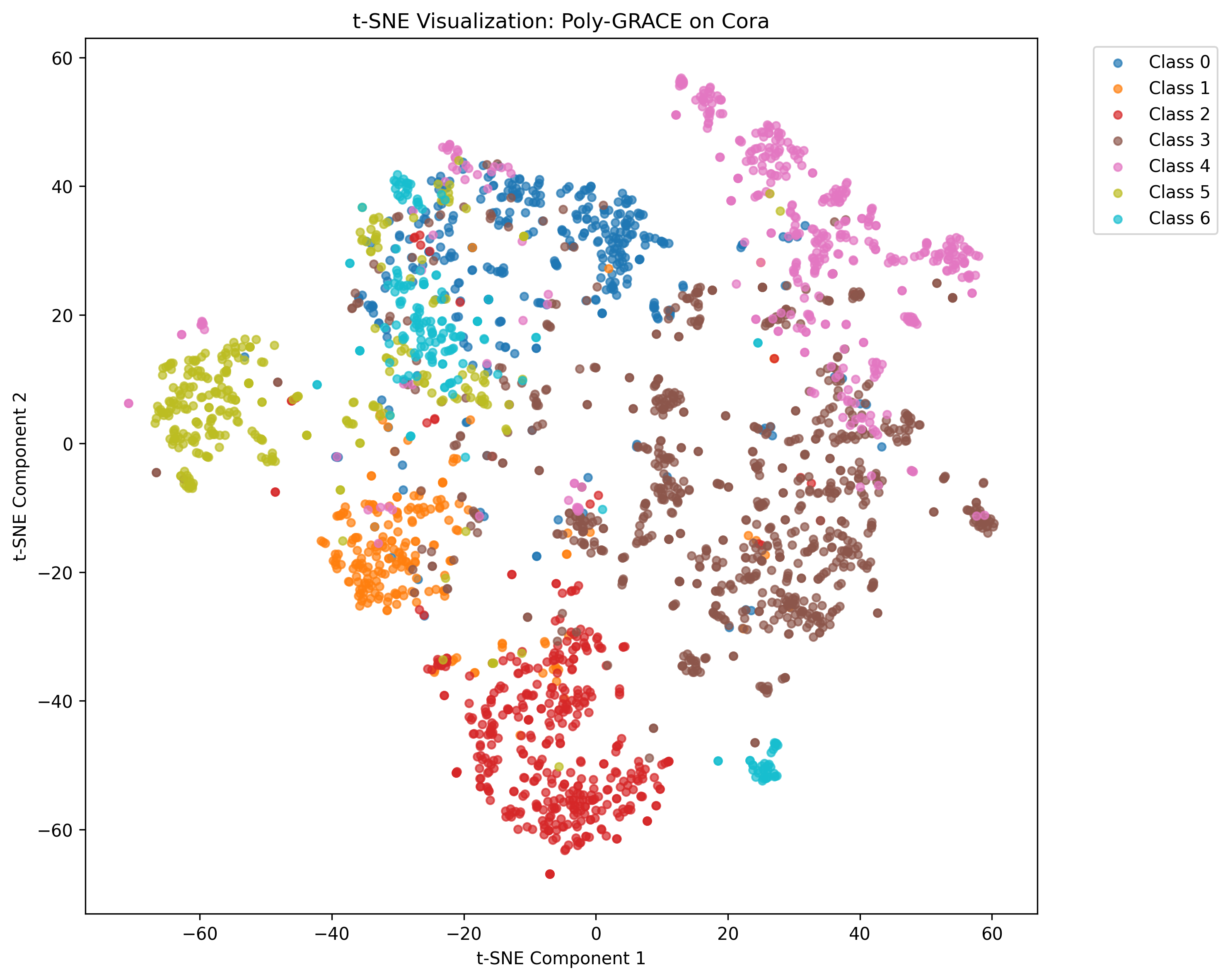}
        \caption{Poly-GRACE (Ours)}
        \label{fig:tsne_polygrace}
    \end{subfigure}
    \caption{t-SNE visualization of node embeddings learned on the Cora dataset. Both methods learn to separate the classes into distinct clusters, demonstrating the effectiveness of the self-supervised pre-training. The quality of the Poly-GRACE embeddings is visually comparable to the non-private baseline.}
    \label{fig:tsne}
\end{figure*}

On the larger and denser PubMed dataset, Poly-GRACE's performance degrades, achieving 51.3\% accuracy compared to GRACE's 79.6\%. This indicates that while our method is highly effective on smaller graphs, its stability and performance on larger graphs may require further hyperparameter tuning or more advanced architectural modifications. The quality of the learned representations is visualized in Fig. \ref{fig:tsne}.

\begin{table}[htbp]
\centering
\caption{Comparison of node classification accuracy (\%) across different datasets.}
\label{tab:main_results}
\begin{tabular}{lccc}
\toprule
\textbf{Dataset} & \textbf{Untrained} & \textbf{GRACE} & \textbf{Poly-GRACE (Ours)} \\
\midrule
Cora & 54.60 & 82.60 & \textbf{79.70} \\
CiteSeer & 48.70 & 66.60 & \textbf{67.80} \\
PubMed & 63.10 & 79.60 & \textbf{51.30} \\
\bottomrule
\end{tabular}
\end{table}

\subsection{Ablation and Sensitivity Analysis}
To better understand the individual contributions of our design choices, we conducted an ablation study on the Cora dataset. The results are presented in Table \ref{tab:ablation_study}. The study reveals that our `poly\_loss` with a standard ReLU-based GCN yields the highest accuracy (82.8\%), demonstrating the power of our novel loss function in isolation. The full Poly-GRACE system maintains a strong performance, confirming that both components contribute effectively and integrate well to form a robust, HE-compatible framework.

\begin{figure}[htbp]
\centering
\includegraphics[width=0.8\columnwidth]{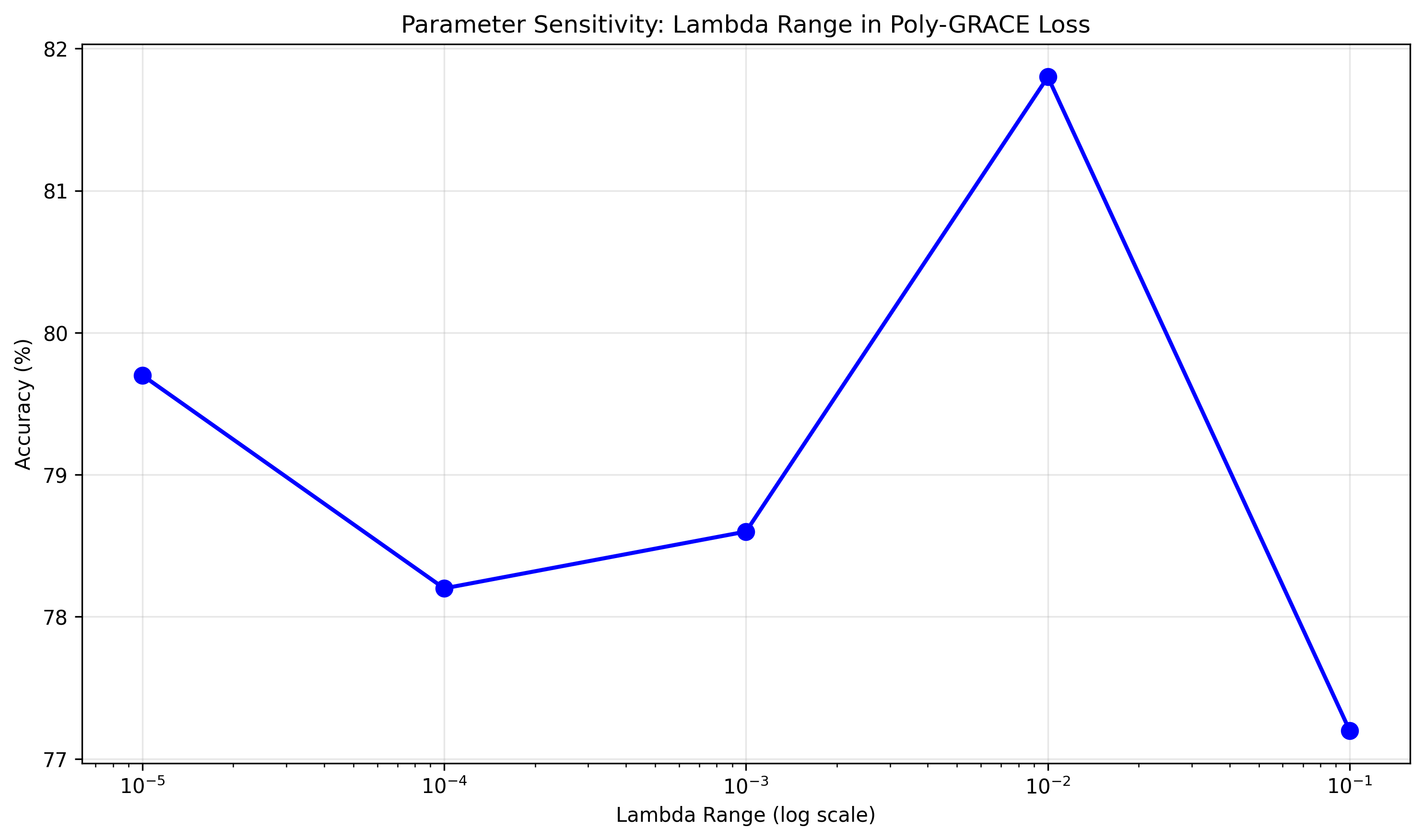}
\caption{Sensitivity analysis of Poly-GRACE's performance on the Cora dataset with respect to the regularization hyperparameter $\lambda$.}
\label{fig:sensitivity}
\end{figure}

Furthermore, we analyzed the sensitivity of our method to the regularization hyperparameter $\lambda$, shown in Fig. \ref{fig:sensitivity}. The performance is stable across several orders of magnitude, with the best result at $\lambda=10^{-2}$, demonstrating that our method is not overly sensitive to this choice.

\begin{table}[htbp]
\centering
\caption{Ablation study on Cora: Impact of activation and loss functions on accuracy (\%).}
\label{tab:ablation_study}
\begin{tabular}{lc}
\toprule
\textbf{Encoder + Loss Combination} & \textbf{Accuracy (\%)} \\
\midrule
GCN (ReLU) + GRACE Loss (Baseline) & 80.80 \\
GCN (ReLU) + Poly Loss & \textbf{82.80} \\
PolyGCN ($x^2$) + GRACE Loss & 81.20 \\
PolyGCN ($x^2$) + Poly Loss (Ours) & 80.80 \\
\bottomrule
\end{tabular}
\end{table}

\section{Limitations and Future Work}
While our Poly-GRACE framework demonstrates promising results, we acknowledge its limitations. Our experiments are focused on transductive node classification tasks on three common citation networks. Further validation is needed to assess the generalizability of our method to other graph types and tasks. Furthermore, the results on the larger PubMed dataset highlight an \textbf{open research challenge in scaling HE-compatible SSL frameworks}. The increased graph size and density likely require more sophisticated hyperparameter tuning or architectural adaptations to maintain stable performance, representing a key direction for future work.

The principles of Poly-GRACE can be extended to more powerful and complex architectures. Future work will focus on adapting our polynomial-friendly framework to Graph Transformers, a task which presents a significant additional challenge: designing a novel, HE-compatible replacement for the non-polynomial Softmax function, a core component of the attention mechanism [8].

\section{Conclusion}
In this paper, we introduced Poly-GRACE, a novel, fully polynomial-friendly framework for self-supervised learning on graphs. By systematically replacing the non-polynomial components in both the GCN encoder and the contrastive loss function, we have created a system that is compatible with Homomorphic Encryption. Our experiments show that it is possible to achieve strong representation learning performance on encrypted graph data, with results that are highly competitive with—and in some cases, exceed—non-private baselines. This work demonstrates that robust privacy guarantees do not have to come at a significant cost to performance and marks a practical step towards enabling large-scale, secure machine learning on sensitive graph-structured data.

\end{document}